\documentclass[10pt,twocolumn,letterpaper]{article}

\usepackage{cvpr}
\usepackage{times}
\usepackage{epsfig}
\usepackage{graphicx}
\usepackage{caption}
\usepackage{subcaption}
\usepackage{amsmath}
\usepackage{amssymb}
\usepackage{mathtools}
\usepackage{color,soul}
\usepackage{array}
\usepackage[titletoc,title]{appendix}

\DeclarePairedDelimiter\floor{\lfloor}{\rfloor}

\usepackage[breaklinks=true,bookmarks=false]{hyperref}

\cvprfinalcopy 

\def\cvprPaperID{****} 

\graphicspath{figure/}
\begin{document}

\title{Video Representation Learning and Latent Concept Mining for Large-scale Multi-label Video Classification}


\author{Po-Yao Huang, Ye Yuan, Zhenzhong Lan, Lu Jiang, Alexander G. Hauptmann \\
School of Computer Science, Carnegie Mellon University\\
{\tt\small \{poyaoh, yey1, lanzhzh, lujiang, alex\}@cs.cmu.edu}
}

\maketitle

\begin{abstract}
We report on CMU Informedia Lab's system used in Google's YouTube 8 Million Video Understanding Challenge. In this multi-label video classification task, our pipeline achieved 84.675\% and 84.662\% GAP on our evaluation split and the official test set. We attribute the good performance to three components: 1) Refined video representation learning with residual links and hypercolumns 2) Latent concept mining which captures interactions among concepts. 3) Learning with temporal segments and weighted multi-model ensemble. We conduct experiments to validate and analyze the contribution of our models. We also share some unsuccessful trials leveraging conventional approaches such as recurrent neural networks for video representation learning for this large-scale video dataset. All the codes to reproduce our results are publicly available\footnote{\url{https://github.com/Martini09/informedia-yt8m-release}}.
\end{abstract}

\section{Introduction}
Ranging from booming personal video collections, surveillance recordings, and professional video documentary archives, we have witnessed an unprecedented growth of a wide range of video data. Numerous methods have been invented to understand video contents and enable searching over huge volumes of accumulated video data. Recently released large-scale video datasets such as Google's YouTube 8 Million (Youtube-8M) video collection bring advancements in video understanding tasks and create new possibilities for many emerging applications such as personalized assistant like Google Home and Microsoft Cortana. Youtube 8M is a~\textit{multi-label video classification} benchmark composed of pre-extracted Inception-v3 features~\cite{szegedy2016rethinking}, labels and their hierarchy in the knowledge graph over more than 8 million videos.


The quantity makes Youtube-8M a unique video classification test-bed. There are 5.7 millions training videos, 1.6 millions validation videos, and 0.8 testing videos respectively. The length of videos range from 120 to 500 seconds. Frame-level features are extracted under 1 frame per second (FPS) sampling rate. Video level features are mean-pooled from frame-level features. The size of topic theme pool is 4,716. Each video is with 3.4 labels on average. In comparison to other weakly-labeled datasets~\cite{jiang2017fcvid}, the precision is reasonably good ($\sim 85\%$) while recall remains poor.

Learning an effective model for video understanding at this scale is challenging for the three reasons:
First, although effort in extracting features at a scale of 8 million is alleviated, the provided frame-level features are prepossessed with some unknown PCA and followed by a simple mean pooling to generate video-level representation. We propose to learn an attentive pooling kernel followed by a refined representation learning module to further boost model performance. 
Second, labels (classes/concepts) are assumed to be independent in the Youtube 8M dataset, which fails to capture the authentic underlying relationship (such as co-occurrence, exclusion and hierarchy) between concepts. We address this issue by learning and incorporating latent concepts for multi-label classification.
Third, multi-model ensemble at this scale is under-explored. We design a systematic model ensemble scheme and quantify the importance over heterogeneous models.


Our contribution in this paper is threefold: 1) We investigate feasible neural architectures to enhance mixture of experts (MoE) model with refined representation learning via residual links and hypercolumns. 2) We introduce a novel latent concept learning layer to capture relationships among concepts 3) We incorporate temporal segment data augmentation and leave-one-out ensemble to further boost classification accuracy.

\section{Related work}
In this section, we briefly discuss the most related previous work falling into two categories: how to train a better individual deep video classification model and how to fuse these individual models to achieve best accuracy. 

At the risk of oversimplification, we identify two classes of methods to improve deep neural networks for video classification. The first class of methods is having more data. Generally, there are two ways that deep models can be benefit from having more data. The first approach is supervised pre-training with external data as in \cite{wang2015towards, wang2016temporal}. This approach is especially useful when the inputs are raw video frames. However, this direction is not feasible for the Youtube-8M dataset since we can only access pre-processed video features. The second approach to have more data is through augmenting internal data ~\cite{wang2015towards}. We design a simple way to augment data by exploiting the fact that video information is highly redundant and the boundary of motion is often arbitrary. In addition to apply augmentation for training, we further extend this approach at the inference phase. 

The second classes of methods is designing better network structure. Likewise, this class of methods can be characterized into two groups: task-independent and task-dependent. There are numerous improvements that are task-independent. For example, dropout \cite{srivastava2014dropout}, inception structure \cite{szegedy2015going}, residual structure \cite{he2016deep}, just name a few. In this work, we explore the residual structure \cite{he2016deep} and its variant \cite{bansal2017pixelnet} and found that these structures not only can help to learn deeper networks but can also improve shallow networks. There are also improvements that are more specific to video classification tasks. Most of these improvements try to capture temporal dependencies among frames. The general lesson is that short-term dependency are useful and easy to capture, but long-terms ones are much more difficult to capture. For example, Simonyan et al. \cite{twostream2014} design a two-stream architecture to capture variances between consecutive frames and there are significant amount of following works that try to further improve the network structures \cite{wang2015towards} or capture longer-term temporal information \cite{diba_tle_2016, wang2016temporal, deep_quantization_16, dovf_lan_2017, VideoLSTM2016}. These work represent the state-of-the-art for video classification and they find that sequence model such as LSTM often perform worse than simple BoF models. These observations are consistent with ours on the Youtube-8M dataset. 

Since the capacity of a single model is limited, researchers often use an ensemble of multiple models to improve video classification accuracy. Multiple models can be learned jointly or separately followed by fusion. MoE~\cite{jacobs1991adaptive} and dropout \cite{srivastava2014dropout} both learn a large number of single models jointly. However, these single models are often need to be homogeneous and the size of the model is also limited. To explore the diverse characteristics of heterogeneous models, we train a set of different MoEs separately and fuse them using leave-one-out~\cite{lan2013cmu} fusion method. Another reason for learning different models is that different modalities of data may be harder to learn at the same time.  
Luckily, we find that concatenating visual data and audio data together and feed them into the same network (early fusion) is better than learning them separately and fuse the results later (late fusion) for the Youtube-8M dataset. 


\begin{figure}
    \centering
    \includegraphics[width=1.0\linewidth]{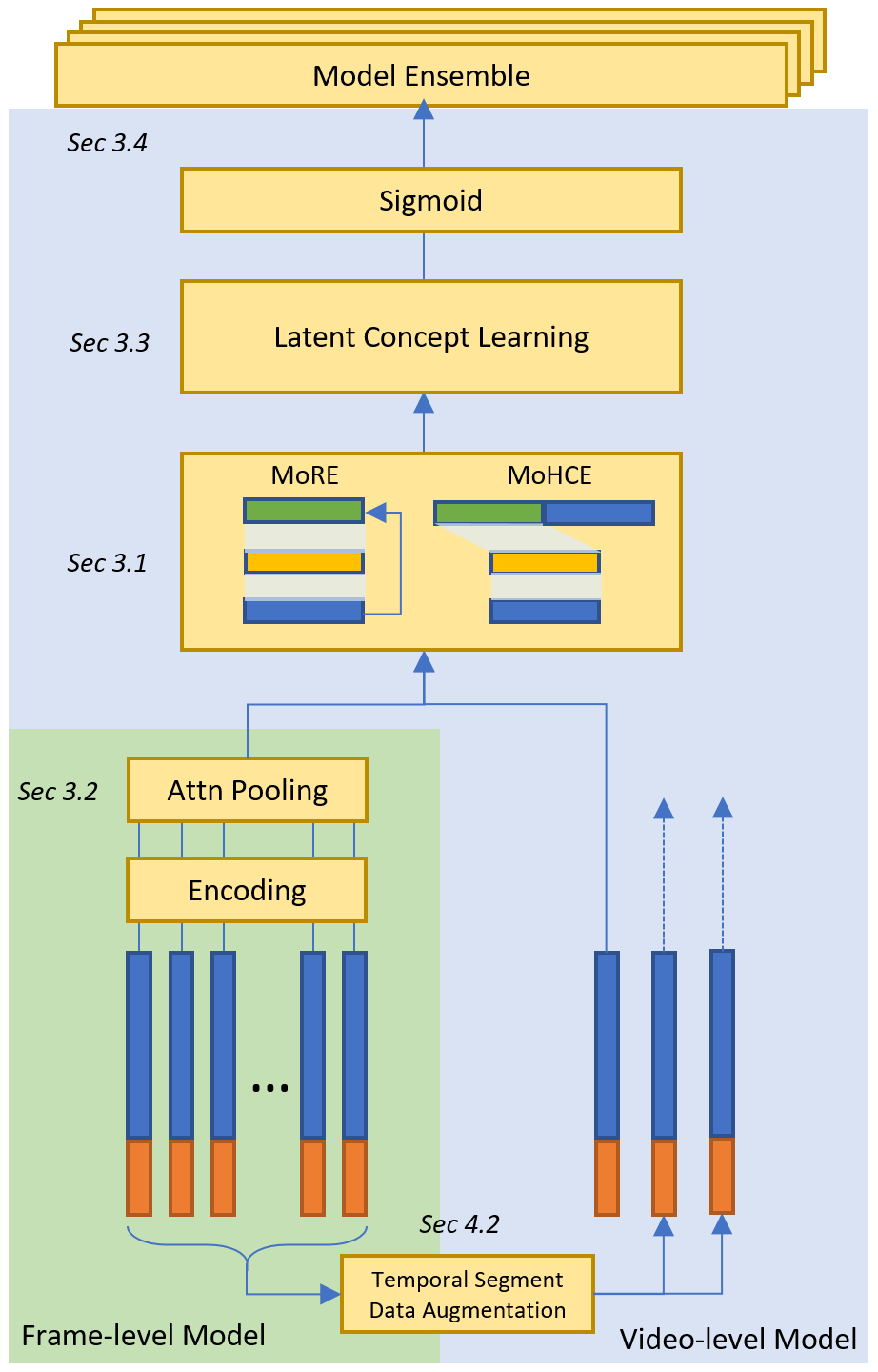}
    \caption{Video-level and Frame-level models}
    \label{fig:all}
\end{figure}

\section{Models} 
Training video-level and frame-level models differ a lot in cost and performance. Without losing choices for practical system design, in this paper, we start with improving video-level models, which is more cost effective, and then we treat frame-level models as models with an additional encoding module that can be stacked upon video-level methods.


\subsection{Video Level Models} 
Built upon MoE models in~\cite{yt8m} without the null experts, we propose to improve the model accuracy in the multi-label video classification task by refining the representation learning with residual links and hypercolumns.

\textbf{Residual Learning}:
While growing MoE deeper, results show that performance gain is mediocre under the same model complexity. As shown in Fig.~\ref{fig:all}, we then apply residual links to the middle layers of expert part of MoE to learn better representation for multi-label classification. We name the combined model Mixture of Residual Experts (MoRE). Residual learning is originally designed to relieve gradient vanishing problem when training very deep neural networks~\cite{he2016deep}. However, we find that simply adding identity mapping to the shallow MoE network also brings significant improvement. Formally, the residual expert network learns the refined representation $\mathbf{x}'$ from the original input $\mathbf{x}$ with:

\begin{equation}
    \mathbf{x}'=F(\mathbf{x},{W_i})+W_s\mathbf{x}
\end{equation}
where in practice $F()$ is stack of a single layer perceptron followed by a batch-normalization layer and dropout layer. We choose a identity matrix for $W_s$. Note that the mixture network which determines the weights of individual classes over experts still take original input instead of the learned representation. We find that using deep structure for the mixture network harm the performance with or without residual links.

\textbf{Hypercolumn}:
Similar to MoRE, we also design Mixture of Hypercolumn Experts (MoHCE). In a typical CNN, higher convolutional layers can capture high-level global context. But they could miss low-level details, thus numerous approaches have built predictors based on the concatenation of multi-stage features \cite{bansal2017pixelnet} \cite{lin2016feature}. We propose to formulate a \textit{hypercolumn} by concatenating features from different layers of a MLP. Formally, the final features are given by:
\begin{equation}
    \mathbf{x'} = [F_1(\mathbf{x}), F_2(\mathbf{x}), ..., F_n(\mathbf{x})]
\end{equation}
where $F_i(\mathbf{x})$ denotes the features of $i$-th layer. This hypercolumn is then fed into mixture of experts (MoE) models to produce final predictions. For MoHCE, we also use standard batch-normalization and dropout techniques to regularize the network training.

\subsection{Frame Level Models}
\textbf{Recurrent Neural Network Models (LSTMs)}. Videos as a collection of frames are inherently rich with temporal information. 
Although many efforts have been made in training recurrent neural networks (RNNs) for video classification, finding a feasible representation remains an open question since its challenging to 1) regularize RNNs and 2) pool representative information frame by frame. Viewing video $v$ as a sequence of frame-level features $x_{1:F_v}^v$, where $x_j^v$ is the features on $j$-th frame, the video-level representation learning from frame-level features with RNN is:

\begin{equation}
\mathbf{x}'=P(\{\Gamma(\mathbf{x}_t,\mathbf{h}_{t-k})\}_{k=1...K,t=1...T}\label{eq:frame_level}
\end{equation}
where $P()$ is the pre-defined or learned pooling kernel. $\Gamma()$ is the RNN function with $t$ as the time index and $k$ as the recurrence time index. For example, in~\cite{yt8m}, the authors use long short term memory (LSTM) network for $\Gamma()$ and the pooling function is a simple concatenation of the last output and the hidden state  $\mathbf{x}=o_{-1}||h_{-1}$. In this work we compare performance of different RNNs and RNNs with the learned pooling kernel by attending to individual frames. On the other hand, since videos usually contain a lengthy sequence of frames, we also experiment regularization methods for RNN such as dropout, zone-out~\cite{krueger2016zoneout}, layer-wise batch-normalization~\cite{cooijmans2016recurrent} to avoid the vanishing gradient problem.   

Intriguingly, experimental results shows that these models with recurrent link do not outperform simple attentive pooling for the PCA-ed features in the Youtube 8M dataset.

\textbf{Attentive Pooling Models (Deep BoF, NetVLAD)}:
Another direction to exploit temporal information is to use attentive pooling of the encoded input features. We propose to remove the RNN module $\Gamma()$ from Eq.~\ref{eq:frame_level} and directly learn a pooling function $P(\{\mathbf{x}_t\})=\sum_t{\alpha_t E(\mathbf{x}_t)}$, where $E(.)$ is some learnable encoding/transformation and $\alpha_t$ is the (attentive) pooling weight. There are many possible design choices of $E(.)$. One example is sparse encoding for deep bag-of-Frames (DBoF) in~\cite{yt8m}, where mean pooling has been utilized to aggregate the encoded frames into video-level representation.

A simple yet effective improvement can be achieved is to make the pooling function learnable. As generalized in NetVLAD~\cite{arandjelovic2016netvlad}, the pooling weights are calculated with a trained soft-attention from frames inputs to some given cluster center. In our frame-level models, we enhance original NetVLAD with an additional multi-layer perceptron for sparse encoding. Formally, the attentive pooling weights are calculated as:  

\begin{equation}
\alpha_{tk}=\frac{\mathbf{w_k}^TE(\mathbf{x}_t)}{\sum_i^K \mathbf{w_k}^TE(\mathbf{x}_i)}
\end{equation}
where $K$ is number of clusters. To make the representation more condense, our model learn another dense layer $G(.)$ to reduce the dimension after concatenation. The final video representation $\mathbf{x}' \in \mathbb{R}^d$becomes:
\begin{equation}
\mathbf{x}'= G(\{\sum_t \alpha_{t1} \{ E(\mathbf{x}_t)- c_1 \} || ... || \sum_t \alpha_{tK} \{ E(\mathbf{x}_t)- c_K \}) 
\end{equation}

\subsection{Latent Concept Learning} 
Classes (or semantic concepts as the more general term) may share complex relationships with each other. As can be seen from the Youtube-8M EDA\footnote{\url{https://www.kaggle.com/philschmidt/youtube8m-eda}}, one concept may frequently co-occur with another concept. For example, chairs and tables can usually be seen together in the indoor-scene videos. Additionally, hierarchical structure of the concepts may also be encoded within the relationships between concepts. For instance, "iPhone6", "iPhone5" belong to the category "iPhone". One approach to disentangle the hierarchical structure among concepts is to map concept names using hand-crafted knowledge graph\footnote{\url{https://developers.google.com/knowledge-graph/}}. Unfortunately, the direct alignments between video labels and the knowledge graph entities remains unclear, making the built-upon graphical structure un-grounded. To address these issues, we propose appending an additional deep neural network at the output of MoRE to directly capture the relationship between concepts. We also show that incorporating these latent concepts would improve the multi-label classification performance.

Unlike the approach in~\cite{sukhbaatar2014training} which uses an additional layer to absorb label noise by twiddling the regularization term, we propose to learn the latent concepts represented in the middle layer of deep residual networks directly and incorporate the original MoRE outputs for final classification. Specifically, we add an additional 2-layered residual network (we called it the latent concept layer, $LC$-layer) with batch normalization as well as input dropout at the output $\mathbf{y}$ of MoRE:
\begin{equation}
\mathbf{y}'= P(G(F(\mathbf{y}, W_{i}^{LC}), W_{s}^{LC})) 
\end{equation}
where $y',y \in \mathbb{R}^{4716}$, $W_{i}^{LC}$ and $W_{s}^{LC}$ are the $LC$-layer parameters, $G()$ is the aggregation function and $P()$ is the final output layer with sigmoid function for multi-label classification. To further alleviate error propagation of ill-classified concepts at the early training stage, we first train without learning latent concepts for 10 epochs and then append a randomly initialize $LC$-layer to start latent concept learning. We find that this late-fire strategy is critical to mine reasonable latent concepts and improve final classification performance. For the aggregation function $G()$, we explore $add()$, $max()$, $append()$ and find that $add()$ delivers the best performance.


\subsection{Ensemble}
We use leave-one-out method~\cite{lan2013cmu} to determine the fusion weights of individual models. For a defined evaluation metric $s$ and model set $M$, the fusion weight $w_m, \sum_m w_m = 1$ for model $m \in M$ is proportional to $d_m = s_{M-m} - s_{M}$, which is the performance drop without model $m$ in comparison to the baseline performance $s_{M}$. A typical choice of the baseline is the result fusing all models equally. In the Youtube-8M video understanding challenge, we use Global Average Precision (GAP) (Details in Section 4) to calculate the weights.

\section{Experiments}
\subsection{Training and Evaluation}

\textbf{Features}: In the Youtube-8M dataset, raw visual features are extracted from Google's Inception-v3 model trained on Imagenet \cite{szegedy2016rethinking}. Raw audio features are extracted from a CNN-inspired architecture trained for audio classification as described in~\cite{hershey2017cnn}. Both visual and audio features follow a PCA whitening process (the PCA matrix is unknown) to further reduce the dimension to 1,024 and 128 respectively. The video-level features are mean-pooled from frame-level features. For training we use the original training set and $\frac{15}{16}$ of the validation set. We evaluated our models on the excluded $\frac{1}{16}$ validation split. The difference in GAP between our $\frac{1}{16}$ validation split and the real test set is less than 0.02\%.

\textbf{Model Training Details}. All our models are trained with cross entropy loss using the Adagrad algorithm~\cite{duchi2011adaptive} with a 0.0002 learning rate and a 0.8 exponential decrease over 8 million examples. We also experimented different loss function such as weighted label loss as in \cite{natarajan2013learning}, optimizer such as momentum and RMSProp but didn't found them beneficial for convergence speed. Common techniques for training neural models including gradient clipping (0.8), dropout (0.8 keep rate), and batch normalization are applied. The mini-batch size is 1,024 (videos) for video level models and 256 (frames) for frame level models.

For video level models, the input is a simple concatenation of l2-normalized mean\_RGB and mean\_audio features in the original dataset. We vary the size of number of mixtures while setting the unit of hidden layer in the residual expert 4,096. Deep MoE models share the same amount of parameter as MoRE but are without residual links. The number of latent concepts we learned in the $LC$-layer is 4,096. The models are trained with 80,000 and 160,000 steps for extended training split and for extended training split with data augmentation (DA) respectively. For frame-level models (DBoF, NetVLAD), we set size of sparse coding 4,096, sample from 80\% of frames, and then compress final representation into a 2,400 dimension vector. The dimension is also 2,400 for LSTM models (dimension for Bi-directional LSTM is doubled).  The typical step for convergence is 360,000.

Our code is based on Tensorflow~\cite{tensorflow}. We use AWS p2-2xlarge instances (Intel Xeon E5-2670 CPU and NVIDIA K80 GPU) to train our models. For most models, we experiment hyper parameters of each model such that each model can maximally fit into 12 GB GPU memory\footnote{MoRE24 and MoHCE12 share roughly the same parameters. MoRE28 and MoRE30 can also fit but the training is sometimes unstable. For frame level models, we use MoRE8 as the classification model}.
There is a notable difference in cost of storage and computation for training video level and frame level models. Training video level models takes around 150 GB storage and around 8 hours to converge while frame-level models takes around 2TB and roughly 5 days to converge. The cost (computation and storage) ratio between a video-level and a frame-level model is roughly 1:16.

\textbf{Evaluation Metric}. We report our results in three metrics: Mean Average Precision (mAP) and Precision at equal recall rate (PERR) \cite{yt8m} and Global Average Precision (GAP). The GAP metric takes the predicted labels that have the highest k (k = 20) confidence scores for each video, then treats each prediction as an individual data point in a long list of global predictions sorted by their confidence scores. The list are then be evaluated with  Average Precision across all of the predictions and all the videos. Formally, 
\begin{equation}
AP=\sum_{i=1}^{N}p(i)\Delta r(i)
\end{equation}
where $N = 20 \times \text{ number if videos}$, $p(i)$ is the precision, and $r(i)$ is the recall given the first $i$ predictions. For detailed definition and interpretation of this new metric, readers may refer to Google's metric page\footnote{\url{https://www.kaggle.com/c/youtube8m##evaluation}}.

\newcolumntype{L}[1]{>{\raggedright\let\newline\\\arraybackslash\hspace{0pt}}m{#1}}
\newcolumntype{P}[1]{>{\centering\arraybackslash}p{#1}}

\subsection{Temporal Segment Data Augmentation}
To alleviate missing of temporal information to boost classification performance while constraining the data size and training time, we propose to temporally segment videos followed by pooling to generate more temporal informative features for training video-level models. Specifically, we temporally split frames in a video in to $N$ segments (each with length $\floor{F_v/s}$) and then perform a fixed pooling functions $P$ with normalization over video segments to generate video-level features: $\{P(x_{s_{i-1}:s_i}^v)\}_{i=1...N}$. In practice we choose $N=3$ and apply a simple mean pooling function $\mu()$ over frame-level RGB and audio features to generate additional video-level features for training. With 4 times larger augmented training data $30\rightarrow120$ GB, we observe consistent performance boost with the training time for convergence will be roughly doubled. Additionally, as can be seen in Appendix, increase of $N$ and other pooling functions such as standard deviation do not help final performance. \cite{wang2016temporal} proposed to randomly sample one short snippet from each segment for training, which in our experiment cause over-fitting in the Youtube-8M dataset.

Similar techniques can also be applied at the inference phase. By feeding the same model with additional features mean-pooled from 3 temporal segments and the original mean-pooled features, we merged the 4 inference results with a fixed weight $(0.1,0.1,0.1,0.7)$ into one final prediction for videos. We name this approach segmented inference.

\subsection{Results of Video-level Models}

\begin{table}[]
\centering
\caption{Video-level Model Performance, LC stands for latent concept learning, SI stands for segmented inference, DA stands for data augmentation}
\label{video_level_peroformance}
\begin{tabular}{lP{0.9cm}P{0.9cm}P{0.9cm}}
Model Name                             & GAP (\%)    & mAP (\%)  & PERR (\%)  \\ \hline \hline
Baseline1 (MoE2)  & 78.41 & 41.58 & 70.9 \\
Baseline3 (MoE8) & 79.30 &  42.20 & 71.9 \\
Deep MoE8 & 79.67 & 43.20  & 72.5 \\
\hline 
MoRE8 & 81.15    & 44.49    &  73.6                     \\
MoRE16 &  81.32    & 44.71    &  73.8                     \\
MoRE24 &  81.61      & 47.30    & 74.3                      \\
MoHCE12 & 82.15 &  48.76   &   74.6                       \\
\hline
MoRE24 + LC & 82.37 & 49.63 & 75.1  \\
MoHCE12 + LC & 82.42  & 47.39  & 74.9 \\
\hline 
MoRE24 + LC + SI & 82.56 & 49.68 & 75.1 \\
MoRE28 + LC + SI & 82.67 &  50.04 & 75.3 \\
MoHCE12 + LC + SI & 82.62 & 47.89 & 75.1 \\
\hline
MoRE28 + LC + DA & 82.78 & 50.47 & 75.5  \\
MoRE30 + LC + DA & 82.79 & 50.74 & 75.6  \\
\hline 
\textbf{MoRE28 + LC + SI + DA} & 82.97 & \textbf{50.89} & 75.6 \\
\textbf{MoHCE12 + LC + SI + DA} & \textbf{83.28} & 50.41 & \textbf{76.0} \\
\hline
\end{tabular}
\end{table}

Table~\ref{video_level_peroformance} summarizes the performance of video-level models. Models with refined representation via residual learning (MoRE) and hypercolumn (MoHCE) deliver a significant $1.5\sim2\%$ gain in GAP on average. Increasing model capacity by adding more experts results in consistent however marginal gain. Learning the latent concept provides an additional $0.5\sim1\%$ performance boost in GAP and PERR. However, we notice that learning latent concept may slightly harm mAP for some models. One possible explanation is the propagation of classification errors from MoRE.

Temporal segment data augmentation has proved to be useful. Data augmentation for training fuels up roughly $0.3\%$ GAP gain. Segmented inference by predicting upon multiple temporal segments brings about $0.2\%$ GAP gain.



\subsection{Results of Frame-level Models}
As can be seen in Table~\ref{frame_level_peroformance}, surprisingly, the recurrent neural network models (with or without recently developed regularization mechanisms) do not achieve good performance. Attentive pooling methods like deep bag-of-feature, VLAD seems to be more feasible for frame-level representation learning. We suspect this phenomenon is rooted from the pre-processed features in the Youtube-8M dataset. Subtle temporal difference may be missing due to PCA. Deep NetVLAD model with smaller codebook would achieve better performance. The decreasing performance with larger codebook size also implies that there might be only a few distinguishable clusters with the PCA-ed features.

Comparing performance of video and frame-level models, the mean-pooled features along with simple data augmentation provide a strong model and achieve competitive single model performance. The result shows that video-level models may be more cost effective than frame-level models.

\begin{table}[]
\centering
\caption{Frame-level Model Performance}
\label{frame_level_peroformance}
\begin{tabular}{lP{1cm}P{1cm}P{1cm}}
Model Name & GAP(\%) & mAP(\%) & PERR(\%) \\ \hline \hline
baseline1 LSTM & 79.43 & 37.43    & 72.2      \\
baseline2 DBoW & 78.34 & 36.97 & 70.9 \\
\hline %
bi-LSTM & 79.98 & 38.51    & 72.5     \\
LSTM + zoneout~\cite{krueger2016zoneout} &  80.15 & 40.27  & 72.4  \\
LSTM + batch\_norm~\cite{cooijmans2016recurrent}  & 80.03  &  40.21 & 72.4 \\
LSTM + attn pooling & 80.10 & 40.33   & 73.2 \\
DBoW + attn pooling & 79.89 & 40.94    &  72.5 \\
VLAD + MoRE8 &  81.71      &    46.18   &   74.1 \\
\hline 
\textbf{NetVLAD + LC ($k=1$)} & \textbf{82.47} & \textbf{48.37} & \textbf{75.1} \\
NetVLAD + LC ($k=2$) & 82.11 & 48.33 & 74.9 \\
NetVLAD + LC ($k=4$) & 81.65 & 47.61 & 74.4 \\
NetVLAD + LC ($k=8$) & 81.29 & 45.54 & 74.1 \\
\hline
\end{tabular}
\end{table}

\subsection{Ensemble Results}
We iteratively grow the ensemble set by randomly adding grouped models from 64 trained models (42 video-level and 22 frame-level models with GAP $> 81.0\%$) followed by a leave-one-out approach to determine fusion weight and remove detrimental models. Table~\ref{ensemble_peroformance} records the growth of set. Both video and frame-level model are required to achieve higher GAP.

Table~\ref{table:ensemble} summarizes the final ensemble set with detailed model weights determined by leave-one-out. Surprisingly, the best single model is not the most important one for ensemble. The individual model weights are not proportional to the original model performances. After reviewing the errors made by different models, we hypothesize that the GAP are greatly affected by hard examples instead of easy examples. The best model may have a wide range of coverage but are not specialized in detecting hard examples. Ensemble of simple models can reach those low-hanging fruits, however, it does not help classifying hard examples. We also found out that regularization mechanisms (e.g. dropout) share the similar phenomena. Training complementary models (for example, ensemble with both video and frame level models) are critical for achieving better GAP. 

We provide our insight why video-level models (with mean-pooled features) and frame-level models can be complimentary. Mean pooling is robust against noises over shots in a video. However, in some cases, a few shots define the labels of a video and therefore frame-level models have a better chance to localize them. Depending on the underlying shot variants of the videos in the Youtube-8M dataset, there might be some trade-off between the two approaches.

Finding a approach to guild training complementary models for ensemble remains open. Readers should be careful when interpreting single model performance and its significance for ensemble.

Our final ensemble achieve 84.68\% GAP on our validation split and 84.66\% on the official test split. Leave-one-out analysis indicates that video level models contributes 54\% of ensemble weight while 46\% from frame-level models. The result shows that to some extent the simple mean-pooled features (video-level models) are strong enough and are crucial for ensemble. Although frame-level models are more general and critical for ensemble, they alone might not be the best choice in sense of performance (roughly the same or slightly worse) and cost ($\sim16:1$) in comparison to video-level models.

\begin{table}[]
\centering
\caption{Model Ensemble Performance. V means number of video-level models and F for frame-level}
\label{ensemble_peroformance}
\begin{tabular}{lllllll}
\# Model & V & F &GAP(\%) & mAP(\%) & PERR(\%) \\ \hline \hline
10 & 10 & 0 & 83.84 & 51.65 & 76.2 \\
12 & 10 & 2 & 84.07 & 52.00 & 76.5 \\
14 & 12 & 2 & 84.34 & 52.86 &  76.8 \\
18 & 12 & 6 & 84.42 & 53.28 & 76.9 \\
18 & 11 & 7 & 84.47 & 53.39 & 76.9 \\
21 & 12 & 9 & 84.52 & 53.64 & 77.0 \\
26 & 17 & 9 & 84.56 & 53.79 & 77.0 \\
\textbf{34} & \textbf{18} & \textbf{16} & \textbf{84.68} & \textbf{54.01} & \textbf{77.2} \\
\hline %
\end{tabular}
\end{table}

\section{Conclusion}
In this paper, we share our exploration of feasible neural network architectures for large-scale video tagging. We introduce two enhanced versions of mixture-of-expert model for video representation learning: mixture-of-residual-expert (MoRE) model and mixture-of-hypercolumn-expert (MoHCE) model that boost performance for both video and frame-level models. 
The proposed delayed-start layer for latent concept learning also demonstrates its favored capability to capture the underlying relationships over concepts and to improve multi-label video classification. The temporal segment data augmentation provides a simple yet effective way to improve video-level models. For frame-level models, we observed that pooling-with-attention methods consistently outperform recurrent neural network models. Ensemble based on leave-one-out  enables us to quantify the importance of heterogeneous models and achieve 84.662\% GAP. While training complementary models seems to be critical, a mechanism to guide complementarity is still under-explored. We would expect a improved framework to learn and ensemble specialized models for large-scale video understanding in the future.

{\small
\bibliographystyle{ieee}
\bibliography{ref}
}

\begin{appendices}

\section{Ensemble Table}


\begin{table*}[t]
\centering
\caption{Final Ensemble Set}
\label{table:ensemble}
\begin{tabular}{lP{1.5cm}P{1.5cm}P{1.5cm}P{2.5cm}}
Model Name                             & GAP (\%)   & mAP (\%) & PERR (\%) & Ensemble Weight \\ \hline \hline
NetVLAD8\_180f\_MoRE8\_2400\_deep\_ckpt274921   &  81.294  &  45.541  &  74.1  &  36  \\
NetVLAD8\_180f\_MoRE8\_2400\_deep\_ckpt280260   &  81.194  &  45.564  &  74.0  &  35 \\  
NetVLAD8\_180f\_MoRE8\_1600\_deep\_ckpt265814   &  81.249  &  46.883  &  74.1  &  34 \\
MoRE30\_LC\_deep\_1st                            &  82.851  &  50.228  &  75.6  &  31 \\
MoRE30\_LC\_deep\_drop\_1st                      &  82.004  &  49.866  &  75.1  &  28 \\
MoRE28\_LC\_deep\_1st                            &  80.427  &  48.214  &  73.7  &  28 \\
MoRE30\_LC\_deep\_2nd                            &  82.868  &  50.522  &  75.6  &  28 \\
NetVLAD4\_180f\_MoRE8\_2400\_deep               &  82.164  &  48.077  &  74.9  &  27 \\
MoRE30\_LC\_deep\_drop\_2nd                      &  82.038  &  50.232  &  75.2  &  27 \\
MoRE30\_LC\_deep\_drop\_3rd                      &  82.786  &  50.744  &  75.6  &  27 \\
MoHCE16\_LC\_1st                                 &  82.702  &  48.593  &  75.3  &  27 \\
NetVLAD2\_180f\_MoRE8\_2400\_deep\_dp           &  82.066  &  48.380  &  74.8  &  26 \\
MoRE28\_LC\_deep\_2nd                            &  81.867  &  48.782  &  74.8  &  26 \\
MoHCE16\_LC\_2nd                                 &  81.881  &  47.614  &  74.6  &  26 \\
NetVLAD8\_180f\_MoRE8                           &  81.827  &  45.308  &  74.3  &  26 \\
MoRE30\_LC\_deep\_3rd                            &  81.734  &  49.663  &  74.8  &  25 \\
MoRE30\_LC\_deep\_l3\_1st                        &  81.104  &  48.789  &  74.3  &  25 \\
MoRE28\_LC\_deep\_3rd                            &  82.783  &  50.472  &  75.5  &  25 \\
MoHCE12\_concat\_deep\_LC                        &  83.281  &  50.412  &  76.0  &  25 \\
MoRE28\_LC\_deep\_4th                            &  82.669  &  50.575  &  75.5  &  25 \\
NetVLAD8\_1200                                  &  82.005  &  45.612  &  74.4  &  25 \\
NetVLAD4\_180f\_MoRE8\_2400\_deep\_dp\_ckpt304488  &  81.650  &  47.577  &  74.5  & 24  \\
MoRE28\_LC\_deep\_5th                            &  81.854  &  50.080  &  74.9  &  24 \\
NetVLAD4\_180f\_MoRE8\_2400\_deep\_dp\_ckpt299007  &  81.604  &  47.610  &  74.4  & 24  \\
NetVLAD1\_180f\_MoRE8\_4096\_deep\_dp           &  82.318  &  48.627  &  75.1  &  24 \\
NetVLAD4\_180f\_MoRE8\_2400                     &  81.971  &  45.607  &  74.5  &  24 \\
MoRE28\_LC\_deep\_6th                            &  81.669  &  49.510  &  74.6  &  23 \\
NetVLAD2\_180f\_MoRE8\_2400\_deep               &  82.102  &  48.281  &  74.8  &  23 \\
MoRE30\_LC\_deep\_l3\_2nd                        &  82.476  &  49.810  &  75.2  &  23 \\
NetVLAD4\_180f\_MoRE8                           &  81.942  &  45.330  &  74.4  &  23 \\
MoRE28\_LC\_residual                             &  82.753  &  50.314  &  75.5  &  22 \\
NetVLAD2\_180f\_MoRE8\_2400\_deep\_dp           &  82.108  &  48.327  &  74.9  &  22 \\
NetVLAD1\_180f\_MoRE8\_4096\_deep               &  82.048  &  48.309  &  74.7  &  22 \\
NetVLAD1\_180f\_MoRE8\_2400\_deep\_dp           &  82.472  &  48.387  &  75.1  &  22 \\
MoRE28\_LC\_deep\_7th                            &  82.700  &  50.016  &  75.4  &  22 \\
 \hline
\end{tabular}
\end{table*}








\end{appendices}

\end{document}